\documentclass[final]{cvpr2021kit/cvpr}  

\usepackage{times}
\usepackage{epsfig}
\usepackage{graphicx}
\usepackage{amsmath}
\usepackage{amssymb}

\usepackage{booktabs}
\usepackage{multirow}
\usepackage{longtable}
\usepackage{colortbl}
\usepackage{tablefootnote}
\usepackage{tabularx}
\usepackage{amsmath}
\usepackage{subcaption}

\usepackage{url}

\usepackage{mathtools}

\usepackage{pbox}
\usepackage[subtle,mathdisplays=normal,indent=normal]{savetrees}

\newcommand{\mparagraph}[1]{\vspace{0.3em}\noindent\textbf{#1}\hspace{1mm}}

\usepackage[pagebackref=true,breaklinks=true,colorlinks,bookmarks=false]{hyperref}

\def\cvprPaperID{4401} 
\def\confYear{CVPR 2021}

\begin{document}

\title{Differentiable Diffusion for Dense Depth Estimation from Multi-view Images} 

\author{Numair Khan\\
Brown University\\
\and
Min H. Kim\\
KAIST\\

\and
James Tompkin\\
Brown University\\
}

\maketitle

\abovedisplayskip=6pt
\abovedisplayshortskip=6pt
\belowdisplayskip=6pt
\belowdisplayshortskip=6pt

\begin{abstract}
\vspace{-0.25cm}
We present a method to estimate dense depth by optimizing a sparse set of points such that their diffusion into a depth map minimizes a multi-view reprojection error from RGB supervision. 
We optimize point positions, depths, and weights with respect to the loss by differential splatting that models points as Gaussians with analytic transmittance. 
Further, we develop an efficient optimization routine that can simultaneously optimize the 50k+ points required for complex scene reconstruction. 
We validate our routine using ground truth data and show high reconstruction quality. 
Then, we apply this to light field and wider baseline images via self supervision, and show improvements in both average and outlier error for depth maps diffused from inaccurate sparse points. 
Finally, we compare qualitative and quantitative results to image processing and deep learning methods.

\end{abstract}

\vspace{-0.40cm}
\section{Introduction}
\label{sec:introduction}
\vspace{-0.1cm}

In multi-view reconstruction problems, estimating dense depth can be difficult for pixels in smooth regions.
As such, 2D diffusion-based techniques~\cite{levin2004, holynski2018} perform gradient-based densification using only a sparse set of depth labels in image space. 
These assume smoothness between points to densify the point set. 
Smoothness can be a good assumption; for instance, many indoor scenes have low texture walls and adhere to the basic assumption that diffusion implies. 
But, diffusion from noisy point samples may produce results with lower accuracy, and it can be difficult to identify and filter out noisy or erroneous points from a sparse set. However, given the correct noise-free constraints, diffusion can be shown to produce comparable or better results than state-of-the-art dense processing methods.

So, how can we handle noisy points? We present a method to optimize point constraints for a set of linear equations representing the solution to the standard Poisson problem of depth diffusion. For this, we develop a differentiable and occlusion-aware image-space representation for a sparse set of scene points that allows us to solve the inverse problem efficiently using gradient descent. We treat each point as a Gaussian to be splatted into the camera, and use the setting of radiative energy transfer through participating media to model the occlusion interaction between Gaussians. This method allows us to optimize over position, depth, and weight parameters per point, and to optimize the point set via reprojection error from multiple RGB images. 

\begin{figure}[t]
\vspace{-3mm}
\includegraphics[width=\linewidth]{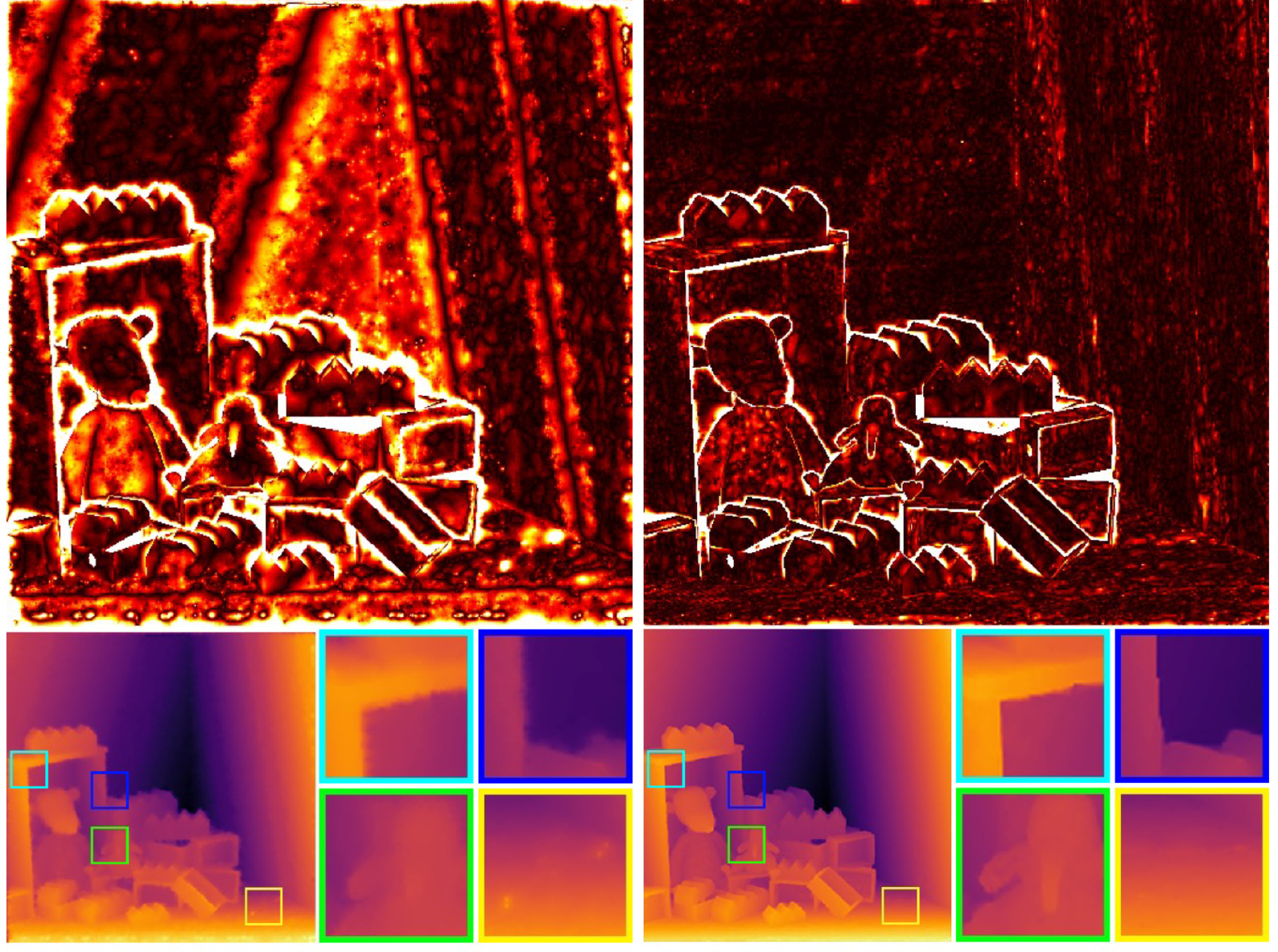}
\vspace{-0.60cm}
\caption[]{\textbf{Left:} Diffusion from an noisy point set produces significant errors from points found along RGB edges instead of depth edges (\emph{top}, log absolute error). In the background, smooth depth regions have banding; in the foreground, RGB texture details are pulled into the depth causing outliers on the floor. \textbf{Right:} Our differentiable point optimization reduces error across the image, removing banding errors and minimizing texture pull.}
\label{fig:teaser}
\vspace{-2.5mm}
\end{figure}

On synthetic and real-world data across narrow-baseline light field multi-view data, and with initial results on wider-baseline unstructured data, we show that our method reduces significant diffusion errors caused by noisy or spurious points. Further, we discuss why edges are difficult to optimize via reprojection from depth maps. Finally, in comparisons to both image processing and deep learning baselines, our method shows competitive performance especially in reducing bad pixels. Put together, we show the promise of direct point optimization for diffusion-based dense depth estimation.

\emph{Data, code, and results:} \href{http://visual.cs.brown.edu/diffdiffdepth}{visual.cs.brown.edu/diffdiffdepth}

\begin{figure*}[t]
\centering
\vspace{-12mm}
\includegraphics[width=\linewidth]{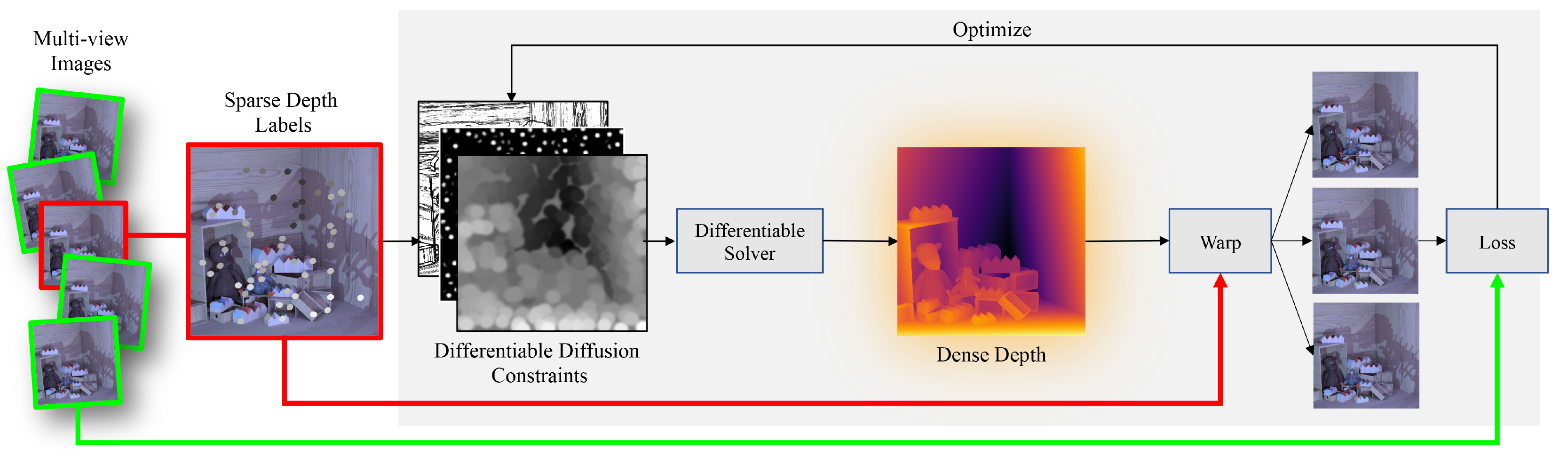}%
\vspace{-4mm}
\caption[]{From a set of noisy sparse depth samples, our method uses differentiable splatting and diffusion to produce a dense depth map. Then, we optimize point position, disparity, and weight against an RGB reprojection loss. This reduces errors in the initial set of points.}
\label{fig:overview}
\vspace{-4mm}
\end{figure*}

\section{Related Work}
\label{sec:related-work}

\vspace{-0.1cm}
\paragraph{Sparse Depth Estimation and Densification}
Our problem begins with sparse depth estimates, e.g., from multiple views~\cite{schoenberger2016sfm}, and relates to dense depth estimation~\cite{schoenberger2016mvs,huang18} and depth densification or completion.
Early work used cross-bilateral filters to complete missing depth samples~\cite{RGBZcamera}.
Chen et al.~learn to upsample low-resolution depth camera input and regularize it from paired RGB data~\cite{chen2018estimating}. 
Imran et al.~consider the problem of depth pixels being interpolated across discontinuities, and compensate by learning inter-depth object mixing~\cite{imran2019depth}.
Efficient computation is also addressed by Holynski and Kopf~\cite{holynski2018}, who estimate disparity maps for augmented reality.
With accurate depth samples, such as from LIDAR, simple image processing-based methods are competitive with more complex learning-based methods~\cite{KuClassicDepthCompletion2018}. We consider the problem of when depth samples themselves may not be accurate, and any resulting densification without correcting the samples will lead to error.

\mparagraph{Depth Estimation for Light Fields}
We demonstrate our results primarily on light fields. Khan et al.~find accurate sparse Epipolar Plane Image (EPI) points using large Prewitt filters~\cite{khan2019}, then diffuses these across all views using occlusion-aware edges~\cite{khan2020}. Zhang et al.~\cite{zhang2016} propose an EPI spinning parallelogram operator with a large support to reliably find points, and To\v{s}i\'{c} and Berkner~\cite{tosic2014} create light field scale-depth spaces with specially adapted convolution kernels. Wang et al.~\cite{wang15,wang16} exploit angular EPI views to address the problem of occlusion, and Tao et al.~\cite{tao2013} uses both correspondence and defocus in a higher-dimensional EPI space for depth estimation. 

Beyond EPIs, Jeon et al.'s~\cite{jeon2015} method exploits defocus and depth too and builds a subpixel cost volume. Chuchwara et al.~\cite{chuchvara2020} present an efficient method based on superpixels and PatchMatch~\cite{barnes09} that works well for wider-baseline views. Chen et al.~\cite{chen2018} estimate occlusion boundaries with superpixels in the central view to regularize depth estimation.

Deep learned `priors' can guide the estimation process. Alperovich et al.~\cite{alperovich2018} showed that an encoder-decoder can be used to perform depth estimation for the central cross-hair of views. Huang et al.'s~\cite{huang18} work can handle an arbitrary number of uncalibrated views. Shin et al.~\cite{shin2018} combine four different angular directions using a CNN, with data augmentation to overcome limited light field training data. Shi et al.~estimate depth by fusing outputs from optical flow networks across a light field~\cite{shi2019}. Jiang et al.~\cite{jiang2018,jiang2019} learn to estimate depth for every pixel in a light field using a low-rank inpainting to complete disoccluded regions. Finally, most recently, Li et al.~use oriented relation networks to learn depth from local EPI analysis~\cite{li2020}.

\mparagraph{Differentiable Rendering}
Unlike the approaches mentioned thus far, we use differentiable rendering to optimize sparse constraints through a differentiable diffusion process. Xu et al.~\cite{xu2019} use differentiable diffusion based on convolutions for coarse depth refinement. We build upon radiative energy transport models that approximate transmittance through a continuously-differentiable isotropic Gaussian representation~\cite{rhodin2015}.
In this area, and related to layered depth images~\cite{ShadeGHS1998}, recent work in differentiable rendering has addressed multi-plane transmittance for view synthesis~\cite{ZhouTFFS2018,Li2020LF}. Other works consider transmittance in voxel-based representations~\cite{LombaSSSLS2019} and for differentiable point cloud rendering~\cite{YifanSWOeS2019,dai2020neural}. A known challenge with differentiable point clouds is backpropagating the 3D point locations through a differentiable renderer via a splatting algorithm~\cite{Richardt2020}. Wiles et al.~\cite{wiles2020} present a neural point cloud renderer that allows gradients to be back propagated to a 3D point cloud for view synthesis. 
We propose a method to meet this challenge by directly rendering depth, and use it to show how to optimize sparse depth samples to correctly reproject RGB samples across multi-view data.

\section{Depth via Differentiable Diffusion}
\label{sec:method}
\vspace{-0.1cm}
Given a set of $n$ multi-view images $\mathcal{I} = \{\mathrm{I}_0, \mathrm{I}_1, ... , \mathrm{I}_n\}$, and a sparse set of noisy scene points $\mathcal{P} \in \mathbb{R}^3$, our goal is to generate a dense depth map for central view $\mathrm{I}_c$. To achieve this, we will optimize the set of scene points such that their diffused image minimizes a reprojection error across $\mathcal{I}$.

\subsection{Depth Diffusion}
\label{sec:depth-diffusion}

\begin{figure*}[t]
\centering
\vspace{-7mm}
\includegraphics[width=\linewidth,clip,trim={2.7cm 6.2cm 0.6cm 0.5cm}]{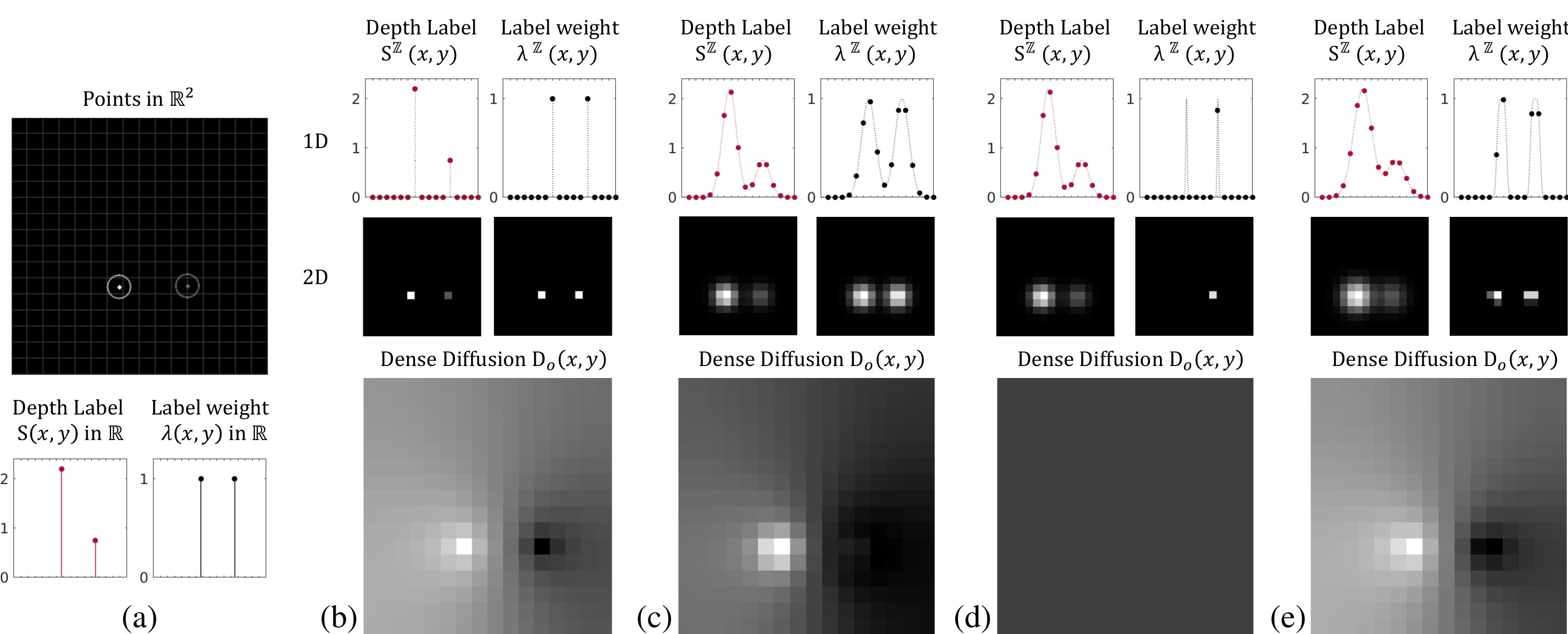}
\vspace{-6mm}
\caption[]{Depth diffusion happens in image space, so how we splat a set of scene points in $\mathbb{R}^3$ onto a pixel grid in $\mathbb{Z}^2$ has a significant impact on the results. \textbf{(a)} The image-space projection of scene points are Dirac delta functions which cannot be represented in discrete pixels. \textbf{(b)} Rounding the projected position to the closest pixel provides the most accurate splatting of depth labels for diffusion, even if it introduces position error. Unfortunately, the functional representation of the splatted point remains a non-differentiable Dirac delta. \textbf{(c)} Image-space Gaussians provide a differentiable representation, but the depth labels are not accurate. Since the label weights $\lambda^\mathbb{Z}$ are no longer point masses, non-zero weight is assigned to off-center depth labels. \textbf{(d)} Attempting to make $\lambda^\mathbb{Z}$ more similar to a point mass by reducing the Gaussian $\sigma$ results in sub-pixel points vanishing: the Gaussian on the left no longer has extent over any of the sampled grid locations. \textbf{(e)} Our higher-order Gaussian representation provides dense diffusion results closest to \textbf{(a)} while also being differentiable.}
\label{fig:splatting}
\vspace{-4mm}
\end{figure*}

\vspace{-0.1cm}
Let $\mathrm{S} \in \mathbb{R}^2$ denote the sparse depth labels obtained by projecting $\mathcal{P}$ onto the image plane of some $\mathrm{I}\in{\mathcal{I}}$. That is, for a given scene point $\mathrm{\textbf{x}} = (X_{\mathrm{\textbf{x}}}, Y_{\mathrm{\textbf{x}}}, Z_{\mathrm{\textbf{x}}}) \in \mathcal{P}$ and camera projection matrix $\mathrm{K}$, $\mathrm{S}(\mathrm{K\textbf{x}}) = Z_{\mathrm{\textbf{x}}}$. We wish to obtain a dense depth map $\mathrm{D}_o$ by penalizing the difference from the sparse labels $\mathrm{S}$ while also promoting smoothness by minimizing the gradient $\nabla \mathrm{D}$:
\begin{align}
\resizebox{0.75\columnwidth}{!}{
	\mbox{\fontsize{10}{12}\selectfont $
	\begin{gathered}
    \mathrm{D}_o =  \underset{\mathrm{D}}{\mathrm{argmin}} \iint_{\Omega} \lambda(x, y) \left(\mathrm{D}(x, y) -  \mathrm{S}(x, y) \right)^2 \\ 
    + \vartheta(x, y) \lVert \nabla \mathrm{D}(x, y) \rVert\,dx\,dy,
    \end{gathered}
$}}
\label{eqn:diffusion-continuous}
\end{align}
where $\lambda(x, y) = \sum_{\mathrm{\textbf{x}}\in\mathcal{P}} \delta((x, y) -\mathrm{K}\mathrm{\textbf{x}} )$ is a sum of point masses centered at the projection of $\mathcal{P}$---the \emph{splatting} function. The second term enforces smoothness; $\vartheta$ is low around depth edges where it is desirable to have high gradients. Solving  Equation~\eqref{eqn:diffusion-continuous} in 3D is expensive and complex, needing,  e.g.,  voxels or a mesh. More practically, the energy in Equation~\eqref{eqn:diffusion-continuous} is minimized over a discrete pixel grid with indices $x,y$:
\begin{align}
\resizebox{0.80\columnwidth}{!}{
	\mbox{\fontsize{10}{12}\selectfont $
	\begin{gathered}
    \mathrm{D}_o =  \underset{\mathrm{D}}{\mathrm{argmin}} \sum_{(x, y)} \bigg( \lambda^\mathbb{Z}(x, y) \left( \mathrm{D}(x, y) -  \mathrm{S^\mathbb{Z}}(x, y) \right)^2   \\ 
     + \sum_{(u, v) \in \mathcal{N}(x, y)}  \vartheta^\mathbb{Z}(x, y)  \lVert \mathrm{D}(u, v) - \mathrm{D}(x, y) \rVert \vphantom{} \bigg),
    \end{gathered}
$}}
\label{eqn:diffusion-discrete}
\end{align}
where $\mathcal{N}(x, y)$ defines a four-pixel neighborhood around  $(x, y)$, and $\lambda^\mathbb{Z}$, $\vartheta^\mathbb{Z}$ and $\mathrm{S}^\mathbb{Z}$ are  respectively the discrete counterparts of the splatting function $\lambda$, the local smoothness weight $\vartheta$, and the depth label in $\mathbb{R}^2$, $\mathrm{S}$. 

Deciding how to perform this discretization has important consequences for the quality of results and is not easy. For instance, $\lambda$ and $\mathrm{S}$ are defined as point masses and hence are impossible to sample. The simplest solution is to round our projected point $\mathrm{K}\mathrm{\textbf{x}}$ to the nearest pixel. However, quite apart from the aliasing that this is liable to cause, it is unsuitable for optimization as the underlying representation of $\lambda^\mathbb{Z}$ and $\mathrm{S}^\mathbb{Z}$ remains non-differentiable. As Figure~\ref{fig:splatting} shows, we require a representation that is differentiable and has the appropriate compactness for correctly representing the weight and depth value of each point on the raster grid: points projected to the raster grid should `spread' their influence only where necessary for differentiability.

\subsection{Differentiable Image-space Representation}
\label{sec:differentiable-representation}
\vspace{-0.1cm}
A common smooth representation is to model the density $\mathrm{\textbf{x}}$ at a three-dimensional scene point as a sum of scaled isotropic Gaussians~\cite{rhodin2015, stoll2011}.
The problem with this approach is that rendering all such points $\mathrm{\textbf{x}}\, \in\, \mathcal{P}$ requires either ray-marching through the scene, or representing the viewing-frustum as a voxel grid. The former is computationally expensive and the latter limits rendering resolution. Moreover, with points defined in scene space, it becomes difficult to ensure depth values are accurately splatted onto discrete pixels. This is demonstrated in Figure~\ref{fig:splatting}(e) where the scene point projecting onto a sub-pixel location ends up with zero pixel weight---effectively vanishing.

\begin{figure}[bt]
\vspace{-6mm}
\includegraphics[width=\linewidth,clip,trim={0cm 0.9cm 0cm 0cm}]{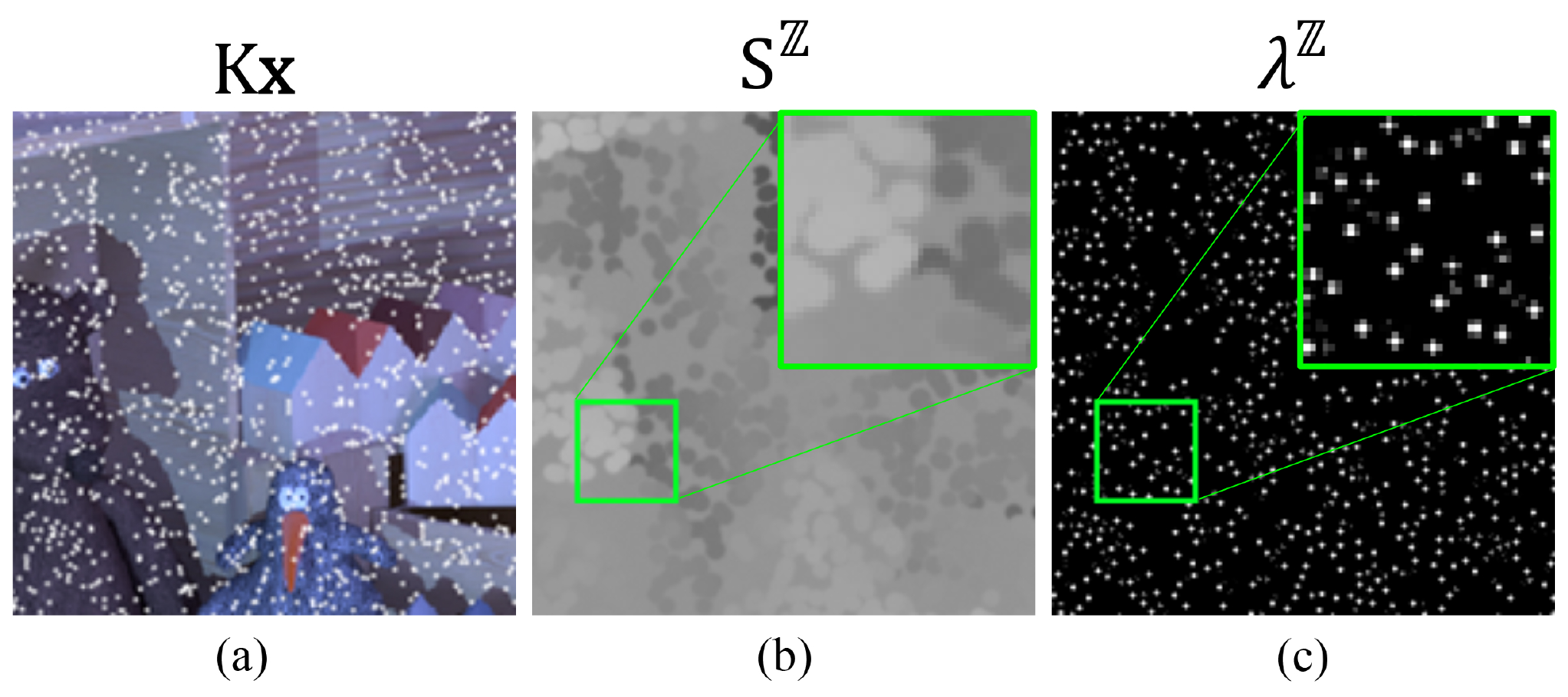}
\vspace{-7mm}
\caption[]{\emph{Left:} The image-space projection $\mathrm{K}\mathrm{\textbf{x}}$ of scene points $\mathrm{\textbf{x}}~\in \mathcal{P}$ plotted in white. \emph{Middle:} Our differentiable labeling function $\mathrm{S}^\mathbb{Z}$ accurately splats depth labels while handling occlusion. \emph{Right:} A higher-order Gaussian representation of $\lambda^\mathbb{Z}$ is differentiable, and provides weights that are close to point masses without any points vanishing during discretization.}
\label{fig:splatted-output}
\vspace{-4mm}
\end{figure}

Our proposed representation overcomes these problems by modeling depth labels as scaled Gaussians centered at the 2D projection $\mathrm{K\textbf{x}}$ of points $\textbf{x}\, \in \,\mathcal{P}$, and using a higher-order Gaussian (or \emph{super-Gaussian}) for the label weight to ensure non-zero pixel contribution from all points. A higher-order Gaussian is useful for representing weight as it has a flatter top, and falls off rapidly. Thus, its behavior is closer to that of a delta function, and it helps minimize the ``leakage'' of weight onto neighboring pixels (Fig.~\ref{fig:splatting}c). But unlike a delta, it is differentiable, and can be sized to match some pixel extent so that points do not vanish (Figs.~\ref{fig:splatted-output}d \&~\ref{fig:splatted-output}e). Thus, we define the discrete functions:
\begin{align}
    \mathrm{S}^\mathbb{Z}(x, y) = \sum_{\mathrm{\textbf{x}}\in\mathcal{P}} \alpha_\mathrm{\textbf{x}}(x, y) \mathrm{S}_{\mathrm{\textbf{x}}}^\mathbb{Z}(x, y),
    \label{eqn:depthlabel-discrete}
\end{align}
where $\alpha_\mathrm{\textbf{x}}(x, y)$ is a function that will merge projected labels in screen space (we will define $\alpha_\mathrm{\textbf{x}}$ in Sec.~\ref{sec:point-rendering}), and $\mathrm{S}_{\mathrm{\textbf{x}}}^{\mathbb{Z}}$ declares the label contribution at pixel $(x, y)$ from a \emph{single} scene point $\mathrm{\textbf{x}} = (X_{\mathrm{\textbf{x}}}, Y_{\mathrm{\textbf{x}}}, Z_{\mathrm{\textbf{x}}})$ with projection $\mathrm{K\textbf{x}} = (x_{\mathrm{\textbf{x}}}, y_{\mathrm{\textbf{x}}})$. We define $\mathrm{S}_{\mathrm{\textbf{x}}}^{\mathbb{Z}}$ as:
\begin{align}
    \mathrm{S}^\mathbb{Z}_\mathrm{\textbf{x}}(x, y) = Z_{\mathrm{\textbf{x}}}\, \mathrm{exp}\bigg( - \frac{(x - x_{\mathrm{\textbf{x}}})^2 + (y - y_{\mathrm{\textbf{x}}})^2}{2\sigma_\mathrm{S}^2} \bigg).
\label{eqn:differentiable-label-representation}
\end{align}
Similarly, the discrete label weights are defined as:
\begin{align}
    \mathrm{\lambda}^\mathbb{Z}(x, y) = \sum_{\mathrm{\textbf{x}}\in\mathcal{P}} \alpha_\mathrm{\textbf{x}}(x, y)\lambda_\mathrm{\textbf{x}}^\mathbb{Z}(x, y) \mathrm,
    \label{eqn:labelweight-discrete}
\end{align}
with $\mathrm{\lambda}^\mathbb{Z}_\mathrm{\textbf{x}}$ taking the higher-order Gaussian form:
\begin{align}
    \mathrm{\lambda}^\mathbb{Z}_\mathrm{\textbf{x}}(x, y) = w_{\mathrm{\textbf{x}}}\mathrm{exp}\bigg( - \frac{(x - x_{\mathrm{\textbf{x}}})^2 + (y - y_{\mathrm{\textbf{x}}})^2}{2\sigma_\lambda^2} \bigg)^p,
    \label{eqn:dataweight-discrete}
\end{align}
for some scaling factor $w_{\mathrm{\textbf{x}}}$.

\mparagraph{Discussion} One might ask why we do not use higher-order Gaussians for the depth label, too. Depth labels require handling occlusion (unlike their weights), and we model this using radiance attenuation in the next section (Sec.~\ref{sec:point-rendering}). Using higher-order Gaussians for depth requires differentiating a transmission integral (upcoming Eq.~\eqref{eqn:attenuation-general}), yet no analytic form exists for higher-order Gaussians (with an isotropic Gaussian, a representation in terms of the \emph{lower} incomplete gamma function $\gamma$ is possible, but the derivative is still notoriously difficult to estimate).
\subsection{Rendering and Occlusion Handling}
\label{sec:point-rendering}
\vspace{-0.1cm}
While a Gaussian has infinite extent, the value of the depth label function $\mathrm{S}^\mathbb{Z}_\mathrm{\textbf{x}}$ and the label weight function $\lambda_\mathrm{\textbf{x}}^\mathbb{Z}$ at non-local pixels will be small and can be safely ignored. However, we need the operator $\alpha_\mathrm{\textbf{x}}$ from Equations~\eqref{eqn:depthlabel-discrete} and~\eqref{eqn:labelweight-discrete} to accumulate values at any local pixel $(x, y)$ that receives significant density contribution from multiple $S_\mathrm{\textbf{x}}^\mathbb{Z}$. 
This accumulation must maintain the differentiability of $\mathrm{S}^\mathbb{Z}$ and must ensure correct occlusion ordering so that an accurate depth label is splatted at $(x, y)$. Using a Z-buffer to handle occlusion by overwriting depth labels and weights from back to front makes $\mathrm{S}^\mathbb{Z}$ non-differentiable. 

\begin{figure}[t]
\centering
\vspace{-6mm}
\includegraphics[width=0.9\linewidth]{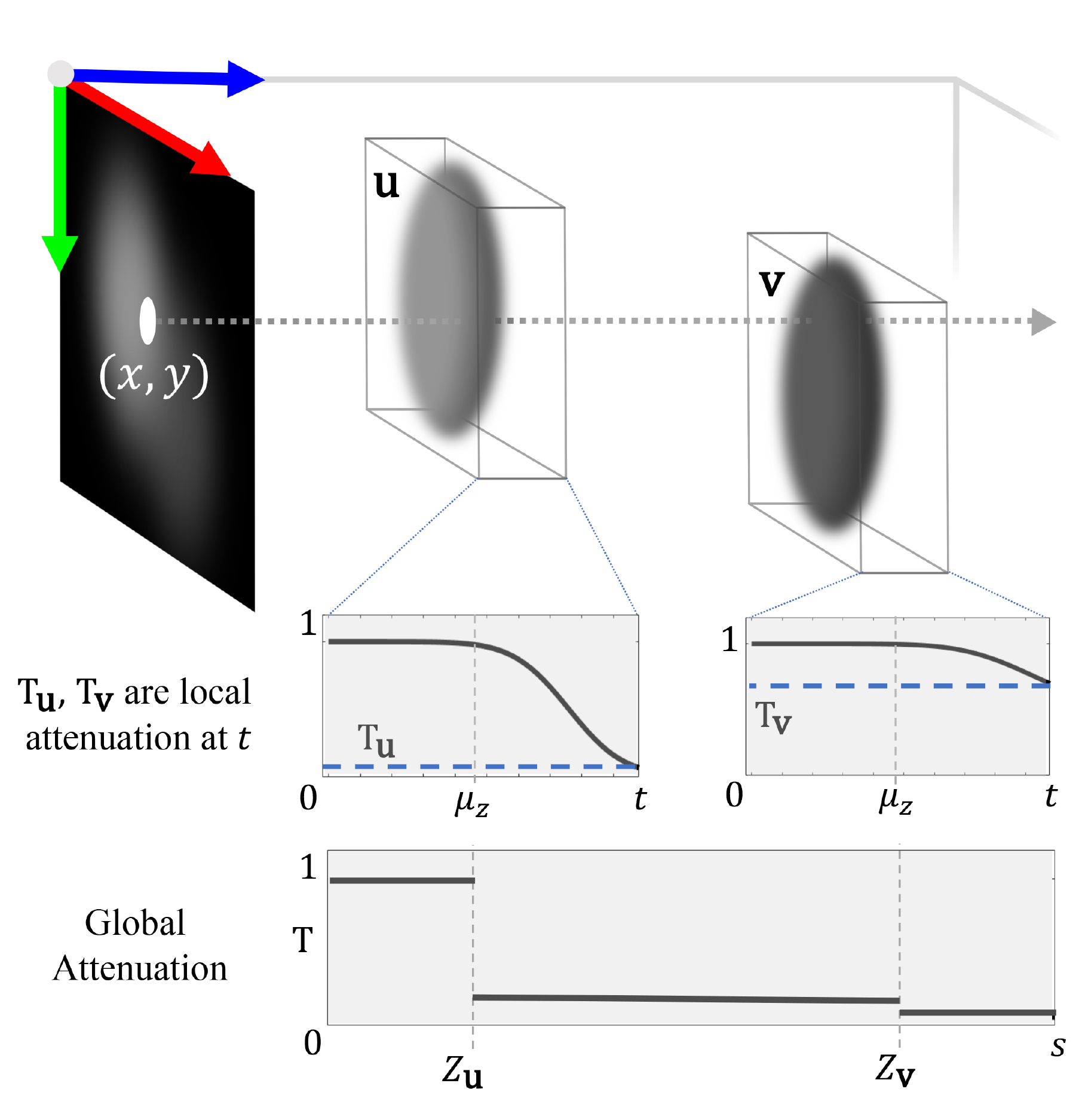}
\vspace{-5mm}
\caption[]{We estimate depth labels at points overlapping in $xy$ using a radiative transfer formulation with Gaussians in orthographic space. If $\sigma_Z$ is small, the influence of the points $\mathrm{\textbf{u}}$ and $\mathrm{\textbf{v}}$ in scene space is restricted to small windows around $Z_\mathrm{\textbf{u}}$ and $Z_\mathrm{\textbf{v}}$. 
As $\sigma_Z\,\xrightarrow{}\,0$, we assume the density contribution at any point $s$ along a ray comes from a single Gaussian. This allows the attenuation effect of each Gaussian to be calculated independently. The global attenuation function at $s$ can be calculated as the product of local attenuation for all points with $z < s$.
}
\label{fig:rendering}
\vspace{-4mm} 
\end{figure}

We diffuse projected points in 2D; however, to motivate and illustrate the derivation of $\alpha_\mathrm{\textbf{x}}$, we will temporarily elevate our differentiable screen-space representation to $\mathbb{R}^3$ and use an orthographic projection---this provides the simplest 3D representation of our `2.5D' data labels, and allows us to formulate $\alpha_\mathrm{\textbf{x}}$ using the tools and settings of radiative energy transfer through participating media~\cite{rhodin2015}.

Thus, we model the density at every 3D scene point as a sum of scaled Gaussians of magnitude $\rho$ centered at the orthographic reprojection $\mathrm{\textbf{u}} = (x_\mathrm{\textbf{x}}, y_\mathrm{\textbf{x}}, Z_\mathrm{\textbf{x}})$ of each $\mathrm{\textbf{x}}~\in~\mathcal{P}$. Then, for a ray originating at pixel $(x, y)$ and traveling along $z$, the attenuation factor $\mathrm{T}$ at distance $s$ from the image plane is defined as:
\begin{align}
\resizebox{0.80\columnwidth}{!}{
	\mbox{\fontsize{10}{12}\selectfont $
	\begin{gathered}
    \mathrm{T}(x, y, s) = \textrm{exp}\bigg(-\int_0^s \rho~ \sum_{\mathrm{\textbf{x}} \in \mathcal{P}} \mathrm{exp}\Bigg(- \Big(\frac{(x - x_\mathrm{\textbf{x}})^2}{2\sigma_\mathrm{S}^2} \\ 
    + \frac{(y - y_\mathrm{\textbf{x}})^2}{2\sigma_\mathrm{S}^2} +  \frac{(z - Z_\mathrm{\textbf{x}})^2}{2\sigma_Z^2}\Big)\Bigg)\,dz \bigg).
    \end{gathered}
$}}
\label{eqn:attenuation-general}
\end{align}
As $\sigma_Z\,\xrightarrow{}\,0$, the density contribution at any point $s$ along the ray will come from only a single Gaussian. Furthermore, as the contribution of each Gaussian is extremely small beyond a certain distance, and as the attenuation along a ray in empty space does not change, we can redefine the bounds of the integral in a local frame of reference. Thus, we consider each Gaussian as centered at $\mu_z$ in its local coordinate frame with non-zero density only on $[0, t]$ (Figure~\ref{fig:rendering}). The independence of Gaussians lets us split the integral over $[0, s]$ into a sum of integrals, each over $[0, t]$ (please see supplemental document for detailed derivation). Using the product rule of exponents, we can rewrite Equation~\eqref{eqn:attenuation-general} as:
\begin{align}
    \mathrm{T}(x, y, s) &= \prod_{\mathrm{\textbf{x}}} \textrm{exp}\Bigg(-\int_0^t \rho\frac{\mathrm{S}_\mathrm{\textbf{x}}^\mathbb{Z}(x, y)}{Z_\mathrm{\textbf{x}}} \mathrm{exp}\bigg(-\frac{(z - \mu_z)^2}{2\sigma_Z^2}\bigg)\,dz \Bigg) \\ \nonumber
    &= \prod_{\mathrm{\textbf{x}}} \mathrm{T}_{\mathrm{\textbf{x}}}(x, y),
\label{eqn:attenuation-parallel}
\end{align}
where the product is over all $\mathrm{\textbf{x}}\in\mathcal{P}~|~Z_\mathrm{\textbf{x}}~<~s$. By looking again at Equation~\eqref{eqn:differentiable-label-representation}, we can see that $\mathrm{S}_\mathrm{\textbf{x}}^\mathbb{Z}(x, y)/Z_\mathrm{\textbf{x}}$ is simply the normalized Gaussian density in $xy$. 

Each $\mathrm{T}_{\mathrm{\textbf{x}}}$ is independent, allowing parallel calculation:
\begin{align}
  \phantom{\mathrm{T}_\mathrm{\textbf{x}}(x, y)}
  &\begin{aligned}
    \mathllap{\mathrm{T}_\mathrm{\textbf{x}}(x, y)} &= \textrm{exp}\Bigg( \sqrt{\frac{\pi}{2}} \frac{\sigma_Z\ \rho\ \mathrm{S}_\mathrm{\textbf{x}}^\mathbb{Z}(x, y)}{Z_\mathrm{\textbf{x}}}\\
      &\qquad \bigg(-\mathrm{erf}\left(\frac{\mu_z}{\sigma_Z\sqrt{2}}\right) - \mathrm{erf}\left(\frac{t - \mu_z}{\sigma_Z\sqrt{2}}\right) \bigg)\Bigg)
  \end{aligned}\\\nonumber
  &\begin{aligned}
    \mathllap{} &= \textrm{exp}\Bigg( c \frac{ \mathrm{S}_\mathrm{\textbf{x}}^\mathbb{Z}(x, y)}{Z_\mathrm{\textbf{x}}}\Bigg),
  \end{aligned}
\end{align}
where $\mathrm{erf}$ is the error function.
We can now define the label contribution of each $\mathrm{\textbf{x}}$ at pixel $(x, y)$. For this, we use the radiative transfer equation which describes the behavior of light passing through a participating medium~\cite{rhodin2015}:
\begin{align}
\resizebox{0.80\columnwidth}{!}{
	\mbox{\fontsize{10}{12}\selectfont $
	\begin{gathered}
    \mathrm{S}^\mathbb{Z}(x, y) = \int_0^\infty \mathrm{T}(s, x, y) \mathrm{a}(s, x, y) \mathrm{P}(s, x, y)\,ds,
\end{gathered} 
$}}
\end{align}
where $\mathrm{T}$, $\mathrm{a}$, and $\mathrm{P}$ are the transmittance, albedo, and density, respectively, at a distance $s$ along a ray originating at $(x, y)$. Albedo represents the proportion of light reflected towards $(x, y)$, and intuitively, we may think of it as the color of the point seen on the image plane in the absence of any occlusion or shadows. In our case, we want the pixel value to be the depth label $Z_\mathrm{\textbf{x}}$. Making this substitution, and plugging in our transmittance and Gaussian density function, we obtain:
\begin{align}
\resizebox{0.80\columnwidth}{!}{
	\mbox{\fontsize{10}{12}\selectfont $
	\begin{gathered}
    \mathrm{S}^\mathbb{Z}(x, y) = \int_0^\infty \mathrm{T}(x, y, s)\sum_{\mathrm{\textbf{x}}\in\mathcal{P}}\,Z_\mathrm{\textbf{x}}\,\rho\,\mathrm{exp}\Bigg(-\frac{(x - x_\mathrm{\textbf{x}})^2}{2\sigma_\mathrm{S}^2}  \\ + \frac{(y - y_\mathrm{\textbf{x}})^2}{2\sigma_\mathrm{S}^2} +  \frac{(s - Z_\mathrm{\textbf{x}})^2}{2\sigma_Z^2}\Bigg)\,ds .
\end{gathered} 
$}}
\end{align}
Again, with $\sigma_Z\,\xrightarrow{}\,0$, the density contribution at a given $s$ may be assumed to come from only a single Gaussian. This lets us remove the summation over $\mathrm{\textbf{x}}$, and estimate the integral by sampling $s$ at step length $\mathrm{d}s$ over a small interval $\mathcal{N}_\mathrm{\textbf{x}}$ around each $Z_\mathrm{\textbf{x}}$:
\begin{align}
  \phantom{ \mathrm{S}^\mathbb{Z}(x, y)}
  &\begin{aligned}
    \mathllap{ \mathrm{S}^\mathbb{Z}(x, y)} &= \sum_{\mathrm{\textbf{x}}\in\mathcal{P}}\sum_{s\in\mathcal{N}_\mathrm{\textbf{x}}} \mathrm{d}s\,\mathrm{T}(x, y, s)\rho\,\mathrm{S}_\mathrm{\textbf{x}}^\mathbb{Z}(x, y) \mathrm{exp}\bigg(-\frac{(s - Z_\mathrm{\textbf{x}})^2}{2\sigma_Z^2}\bigg) \\ \nonumber
  \end{aligned}\\[-4mm]
  &\begin{aligned}
    \mathllap{} &= \sum_{\mathrm{\textbf{x}}\in\mathcal{P}} \mathrm{S}_\mathrm{\textbf{x}}^\mathbb{Z}(x, y)
    \sum_{s\in\mathcal{N}_\mathrm{\textbf{x}}} \mathrm{d}s\,\mathrm{T}(x, y, s)~\rho~ \mathrm{exp}\bigg(-\frac{(s - Z_\mathrm{\textbf{x}})^2}{2\sigma_Z^2}\bigg) \\ \nonumber
  \end{aligned}\\[-4mm]
  &\begin{aligned}
    \mathllap{} &= \sum_{\mathrm{\textbf{x}}\in\mathcal{P}} \alpha_\mathrm{\textbf{x}}(x, y) \mathrm{S}_\mathrm{\textbf{x}}^\mathbb{Z}(x, y).
  \end{aligned}
\end{align}
This allows us to arrive at a differentiable form of our screen-space aggregation function $\alpha_\mathbf{x}$:
\begin{align}
    \alpha_\mathrm{\textbf{x}}(x, y) &=  \frac{\rho\,\mathrm{d}s}{Z_\mathrm{\textbf{x}}}
    \sum_{s\in\mathcal{N}_\mathrm{\textbf{x}}} \mathrm{T}(s, x, y)~\rho~ \mathrm{exp}\bigg(-\frac{(s - Z_\mathrm{\textbf{x}})^2}{2\sigma_Z^2}\bigg).
\label{eqn:alpha}
\end{align}

\subsection{Optimization by Gradient Descent}
\label{sec:optimization}
\vspace{-0.1cm}
To restate our goal, we want to optimize the parameters $\Theta = \{ \mathrm{S}^\mathbb{Z}, \lambda^\mathbb{Z}, \vartheta^\mathbb{Z}\}$ for dense depth diffusion~(Eq.~\eqref{eqn:diffusion-discrete}). The function $\mathrm{S}^\mathbb{Z}(x, y)$ proposes a depth label at pixel $(x, y)$, $\lambda^\mathbb{Z}(x, y)$ determines how strictly this label is applied to the pixel, and $\vartheta^\mathbb{Z}(x, y)$ controls the smoothness of the output depth map at $(x, y)$. We find $\Theta$ by using gradient descent to minimize a loss function $\mathrm{L}(\Theta)$. Using our differentiable representation, we can express $\mathrm{S}^\mathbb{Z}$ and $\lambda^\mathbb{Z}$ in terms of the image-space projection of the sparse point set $\mathcal{P}$. Doing so provides strong constraints on both the initial value of these two functions, and on how they are updated at each step of the optimization, leading to faster convergence. 

\mparagraph{Supervised Loss} 
To validate our image-space representation and optimization, we first use ground truth depth to supervise the optimization of the different parameters in $\Theta$. This is effective and generates high-quality depth maps; we refer the reader to the supplemental document for details. This shows the potential of our differentiable sparse point optimization and diffusion method, and inform us of the contribution on the final result of the follow self-supervised loss for captured images. 

\mparagraph{Self-supervised Loss} 
Working with a set of multi-view images $\mathcal{I} = \{\mathrm{I}_0, \mathrm{I}_1, ... , \mathrm{I}_n\}$ allows us to define a self-supervised loss function for the optimization. Given a dense depth map $\mathrm{D}_\Theta$ generated by diffusion with parameters $\Theta$, we define the warping operator $\mathcal{W}_\Theta$ to reproject each view $\mathrm{I}^i$ onto $\mathrm{I_c}$; where $\mathrm{I_c}$ is the view we want to compute dense depth for. The warping error is then calculated as:
\begin{align}
\resizebox{\columnwidth}{!}{
	\mbox{\fontsize{10}{12}\selectfont $
	\begin{gathered}
    \mathrm{E}_\Theta(x, y) =  \frac{1}{\sum_i \mathrm{M}^i_\Theta (x, y) + \epsilon} \sum_i \bigg( \lvert \mathrm{I}(x, y) -  \mathcal{W}_\Theta[\mathrm{I}^i](x, y) \rvert\ \mathrm{M}^i_\Theta(x, y) \bigg),
    \end{gathered}
$}}
\label{eqn:error-reprojection}  
\end{align}
where $\mathrm{M}^i_\Theta(x, y)$ is the binary occlusion mask for view $i$, computed dynamically at each iteration. 

\begin{figure}[bt]
\centering
\includegraphics[width=1.0\linewidth]{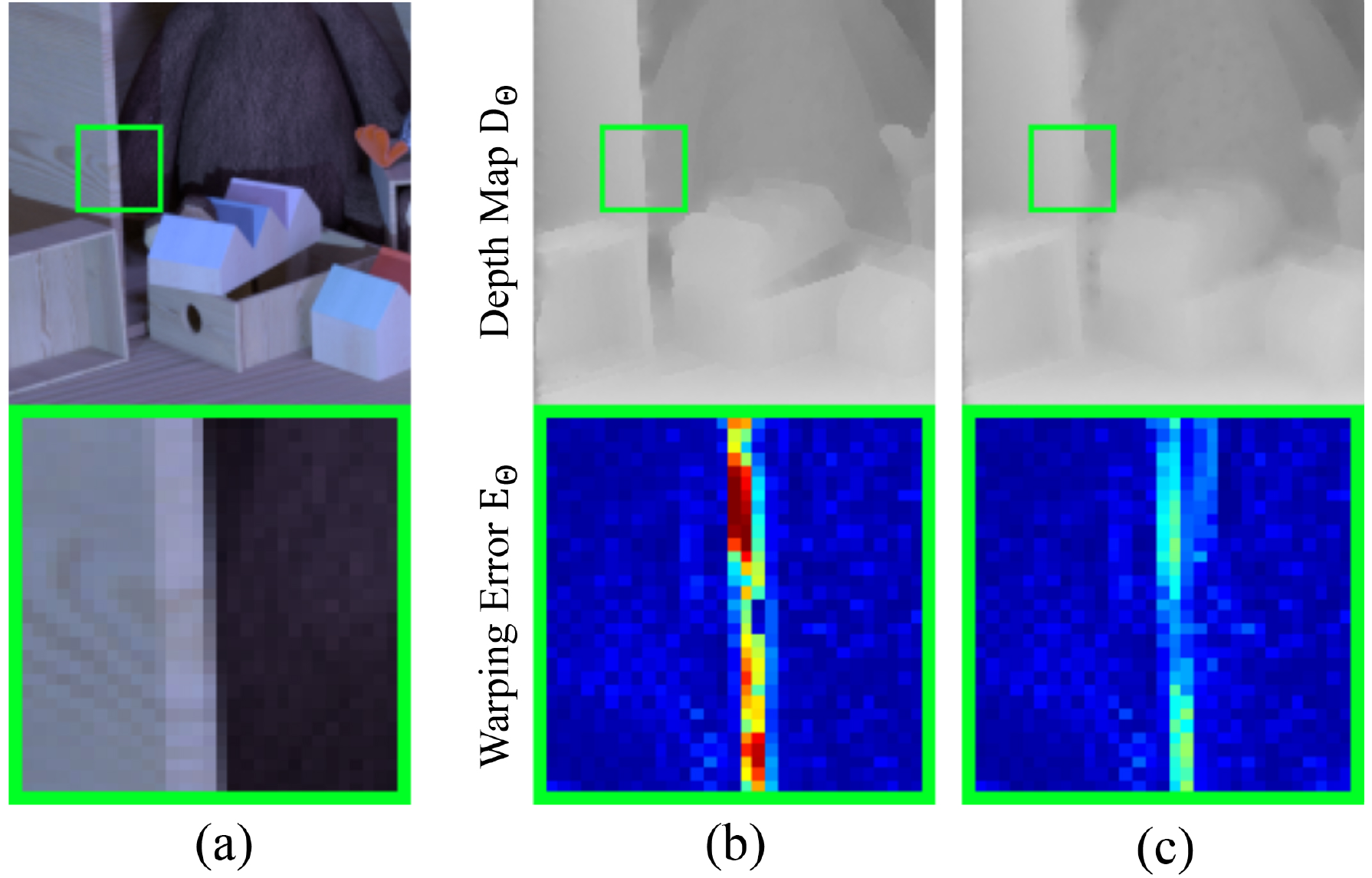}%
\vspace{-4mm}
\caption[]{Our everyday intuition says that depth edges should be sharp, but a limited sampling rate blurs them in the RGB input~\textbf{(a)}. This can cause unintended high error during optimization via losses computed on RGB reprojections. In \textbf{(b)}, the depth edge is sharp, but reprojecting it into other views via warping causes high error as the edge in the RGB image is blurred. Counterintuitively, in \textbf{(c)}, the depth edge is soft and less accurate, but leads to a lower reprojection error. If sharp edges are desired, we can reward high gradient edges in the error (Eq.~\eqref{eqn:losswgradient}).
}
\label{fig:error}
\vspace{-2mm}
\end{figure}


\begin{figure}[bt]
\vspace{0mm}
\includegraphics[width=\linewidth]{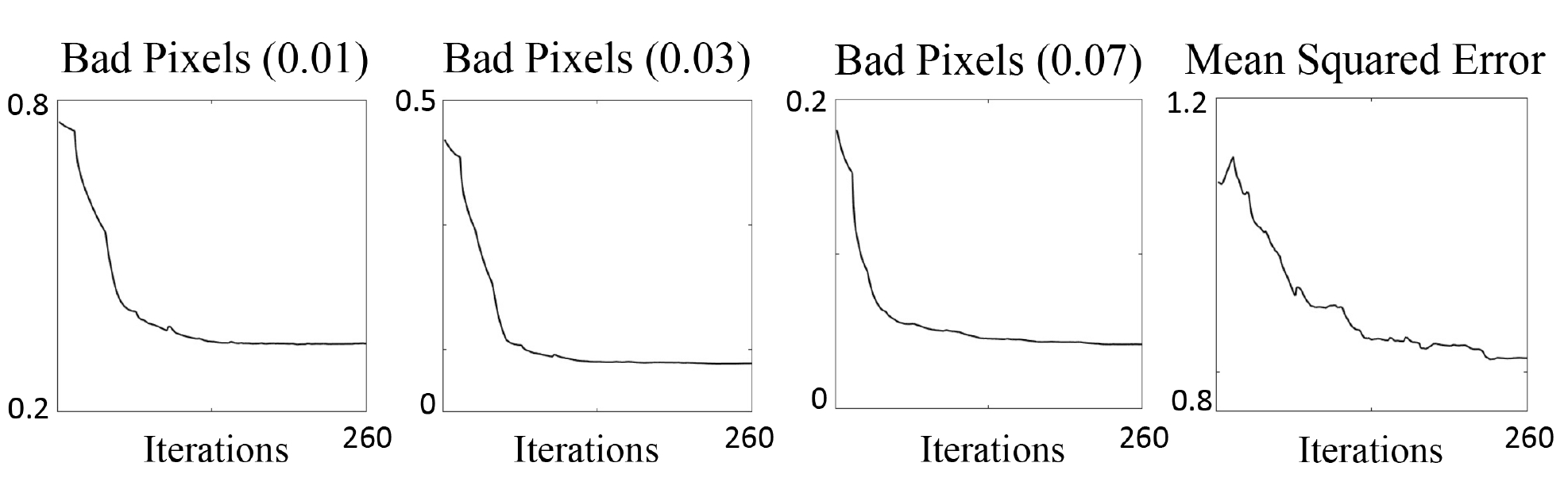}%
\vspace{-3mm}
\caption[]{Over optimization iterations, mean squared error reduces and `bad pixels' are significantly suppressed. From the corresponding error map in Figure~\ref{fig:teaser} we can see that most remaining errors lie along edges, where depth is not well defined (as per Figure~\ref{fig:error})---the ground truth depth values are pixel-rounded while the RGB image contains blurred depth edges.}
\label{fig:delta}
\vspace{-2mm}
\end{figure}

We observe that $\mathrm{E_\Theta}$ is non-zero even if we use the ground truth depth map, because small pixel errors are inevitable during the sub-pixel interpolation for warping. 
However, the more significant errors come from an unexpected source: the sharpness of depth edges.
Depth labels are ambiguous at pixels lying on RGB edges, and limited sampling frequency blurs these edges within pixels~(Figure~\ref{fig:error}). 
By assigning a fixed label to these pixels, sharp depth edges cause large errors. 
Consequently, the optimization process smooths all edges. While doing so minimizes the reprojection error, it may be desirable to have sharp depth edges for aesthetic and practical purposes, even if the edge location is slightly incorrect. 

Therefore, we add a loss term to reward high gradients in $\mathrm{E}_\Theta$, effectively allowing the optimization to ignore errors caused by sharp depth edges. In addition, we include a smoothness term $\mathrm{E}_\mathrm{S}$ similar to Ranjan et al.~\cite{ranjan2019} to encourage depth to be guided by image edges, and a structural self-similarity error~\cite{wang2004} $\mathrm{E}_{\mathrm{SSIM}}$ which is commonly used to regularize warping error. Our final loss function becomes:
\begin{align}
\mathrm{L}(\Theta) = \sum_{(x, y)} \bigg(\mathrm{E}_\Theta(x, y) + \mathrm{E}_\mathrm{S} + \mathrm{E}_{\mathrm{SSIM}} - \nabla \mathrm{E}_\Theta(x, y) \bigg).
\label{eqn:losswgradient}  
\end{align}

\subsection{Implementation}
\label{sec:implementation}

Our proposed framing of the diffusion problem allows us to express $\mathrm{S}^\mathbb{Z}$ and $\lambda^\mathbb{Z}$ as differentiable functions of the points set $\mathcal{P}$, and thus, to calculate $\partial\,\mathrm{L}/\partial\,\mathrm{\textbf{x}}$. 
Since $\mathcal{P}$ provides strong constraints on the shape of these functions, we optimize over the parameters $\mathrm{Z_\textbf{x}}$, $\mathrm{K}\textbf{x}$, $w_\mathrm{\textbf{x}}$, and $\vartheta^\mathbb{Z}$ instead of directly over $\Theta$ ($w_\mathrm{\textbf{x}}$ is the scaling factor from Equation~\eqref{eqn:dataweight-discrete}). 
To regularize the smoothness and data weights, we further define $\vartheta^\mathbb{Z}(x, y) = \mathrm{exp}(-Q(x, y))$ and $w_\mathrm{\textbf{x}} = \mathrm{exp}(-R(\textbf{x}))$, for some unconstrained $R$ and $Q$ that are optimized. Thus, our final parameter set is $\bar\Theta = \{ \mathrm{Z_\mathrm{\textbf{x}}, \mathrm{K}\textbf{x}}, R, Q \}$. We initialize $R(\mathrm{\textbf{x}})$ to zero for all $\textbf{x}$, and $Q(x, y)$ to the magnitude of the image gradient $\lVert \nabla I \rVert$. 

\mparagraph{Distance}
Both RGB and VGG16 features can be used as distances for warping loss $\mathrm{E}_\Theta$; we found VGG16 features to outperform RGB. VGG loss has a better notion of space from a larger receptive field and handles textureless regions better. Thus, we take each warped image in Equation~\eqref{eqn:error-reprojection}, run a forward pass through VGG16, then compute an $L_1$ distance between the 64 convolution activation maps of the first two layers. $\nabla \mathrm{E}_\mathrm{\Theta}$ is computed using the 2D channel-wise mean of $\mathrm{E}_\Theta$; $\mathrm{E}_\mathrm{S}$ and $\mathrm{E}_\mathrm{SSIM}$ are calculated in RGB space.

\mparagraph{Hyperparameters}
This require a trade-off between resource use and accuracy. The parameter $\sigma_\mathrm{S}$ in Equation~\eqref{eqn:depthlabel-discrete} determines the pixel area of a splatted depth label $Z_\mathrm{\textbf{x}}$. Ideally, we want the label to be $Z_\mathrm{\textbf{x}}$ over all pixels where $\lambda_\mathrm{\textbf{x}}^\mathbb{Z} > \epsilon$. The case where the label falls off while the weight is much larger than zero is illustrated in Figure~\ref{fig:splatting}(c) and  leads to incorrect diffusion results. However, ensuring a uniform weight requires having a large value of $\sigma_\mathrm{S}$, and this may cause the labels of neighboring points to be occluded. We found that using $\sigma_\mathrm{S}=~$1.3 provides a good balance between accuracy and compactness. This spreads the label density over three pixels in each direction before it vanishes, so we use a Gaussian kernel size of 7$\times$7. 

For $\sigma_Z$, we want the spread to be as small as possible. However, if the value is very small then we must use a large number of samples in $\mathcal{N}_\mathrm{\textbf{x}}$ when calculating the quadrature in Equation~\eqref{eqn:alpha}. An insufficient number of samples causes aliasing when calculating $\alpha_\mathrm{\textbf{x}}$ at different pixel locations $(x, y)$. A value of $\sigma_Z =~$1.0 and 8 samples in each $\mathcal{N}_\mathrm{\textbf{x}}$ works well in practice.

We use a Gaussian of order $p=~$2 to represent $\lambda_\mathrm{\textbf{x}}^\mathbb{Z}$ (Eq.~\eqref{eqn:labelweight-discrete}). As the order is increased, the Gaussian becomes more similar to a box function and leaks less weight onto neighboring pixels. However, its gradients become smaller, and the loss takes longer to converge. With $p=~$2, we calculate $\sigma_\lambda=~$0.71 to provide the necessary density to prevent points from vanishing (Fig.~\ref{fig:splatting}(d)).

\mparagraph{Routine}
We use Adam~\cite{kingma2014}. We observe a lower loss when a single parameter is optimized at once. Thus, we optimize each parameter separately for 13 iterations, and repeat for 5 passes.

%

\begin{figure*}[tp]
\centering
\includegraphics[width=\linewidth]{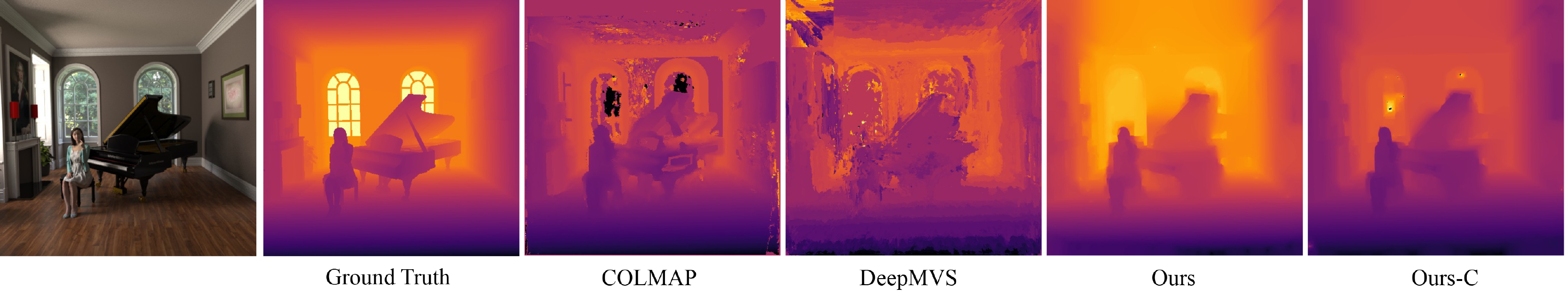}
\vspace{-0.6cm}
\caption{Depth results on the synthetic \emph{Piano-MVS} scene. \emph{Left to right:} Ground truth, dense reconstruction from COLMAP~\cite{schoenberger2016mvs, schoenberger2016sfm}, DeepMVS~\cite{huang18}, our method using $702$ sparse points in $\mathcal{P}$, and our method with $2,808$ sparse points from dense COLMAP output.}
\label{fig:results-mvs}
\vspace{-2.5mm}
\end{figure*}

\mparagraph{Efficiency}
The set of edge pixels require to represent a high-resolution image can run into the tens of thousands, and naively optimizing for this many points is expensive. This is true both of computation time and of memory. Calculating $\mathrm{S}^\mathbb{Z}$ in Equation~\eqref{eqn:depthlabel-discrete} by summing over all points $\mathrm{\textbf{x}}$ is impossibly slow for any scene of reasonable complexity. Fortunately, in practice we only need to sum the contribution from a few points $\mathrm{\textbf{x}}$ at each pixel and, so, the computation of $\alpha_{\mathrm{\textbf{x}}}(x, y)$ in Equation~\eqref{eqn:alpha} is serialized by depth only for points in a local neighborhood. By splitting the image plane into overlapping tiles, non-local points $\mathrm{K}\mathrm{\textbf{x}}$ can be rendered in parallel. The amount of overlap equals the kernel size in $xy$, and is needed to account for points that may lie close to the boundary in neighboring tiles. Using this parallelization scheme, we can render more than 50k points in correct depth order, solve the diffusion problem of Equation~\eqref{eqn:diffusion-discrete}, and back-propagate gradients through the solver and renderer in five seconds.

\mparagraph{Software and Hardware}
\label{sec:software}
We implement our method in PyTorch. For diffusion, we implement a differentiable version of Szeliski's LAHBPCG solver \cite{szeliski2006}. 
All CPU code was run on an AMD Ryzen Threadripper 2950X 16-Core Processor, and GPU code on an NVIDIA GeForce RTX 2080Ti.

\section{Experiments}
\label{sec:experiments}
\label{sec:setting}
\label{sec:results}

\subsection{Light Fields}

\mparagraph{Datasets}
Along with the \textit{Dino}, \textit{Sideboard}, \textit{Cotton}, and \textit{Boxes} scenes from the synthetic HCI Dataset~\cite{honauer2016}, we add two new \textit{Living Room} and \textit{Piano} scenes with more realistic lighting, materials, and depth ranges. We path trace these with Arnold in Maya. All synthetic light fields have 9$\times$9 views, each of 512 $\times$ 512 pixels. 
For real-world scenes, we use light fields from the New Stanford Light Field Archive~\cite{stanford2008}. These light fields have 17$\times$17 views captured from a camera rig, with a wider baseline and high spatial resolution (we downsample 2$\times$ for memory).

\mparagraph{Baselines and Metrics}
For our method, we use an initial point set extracted from EPI edge filters~\cite{khan2019}. We compare to the methods of Zhang et al.~\cite{zhang2016}, Khan et al.~\cite{khan2020}, Jiang et al.~\cite{jiang2018}, Shi et al.~\cite{shi2019}, and Li et al.~\cite{li2020}. Khan et al.'s algorithm is diffusion-based, whereas the last three methods are deep-learning-based. 

%
For metrics, we use mean-squared error (MSE) and \textit{bad pixels} (BP). BP measures the percentage of pixels with error higher than a threshold. For real-world scenes without ground truth depth, we provide a measure of performance as the reprojection error in LAB induced by depth-warping the central view onto the corner views; please see our supplemental material.

\begin{figure*}[htbp]
\centering
\includegraphics[width=0.95\linewidth]{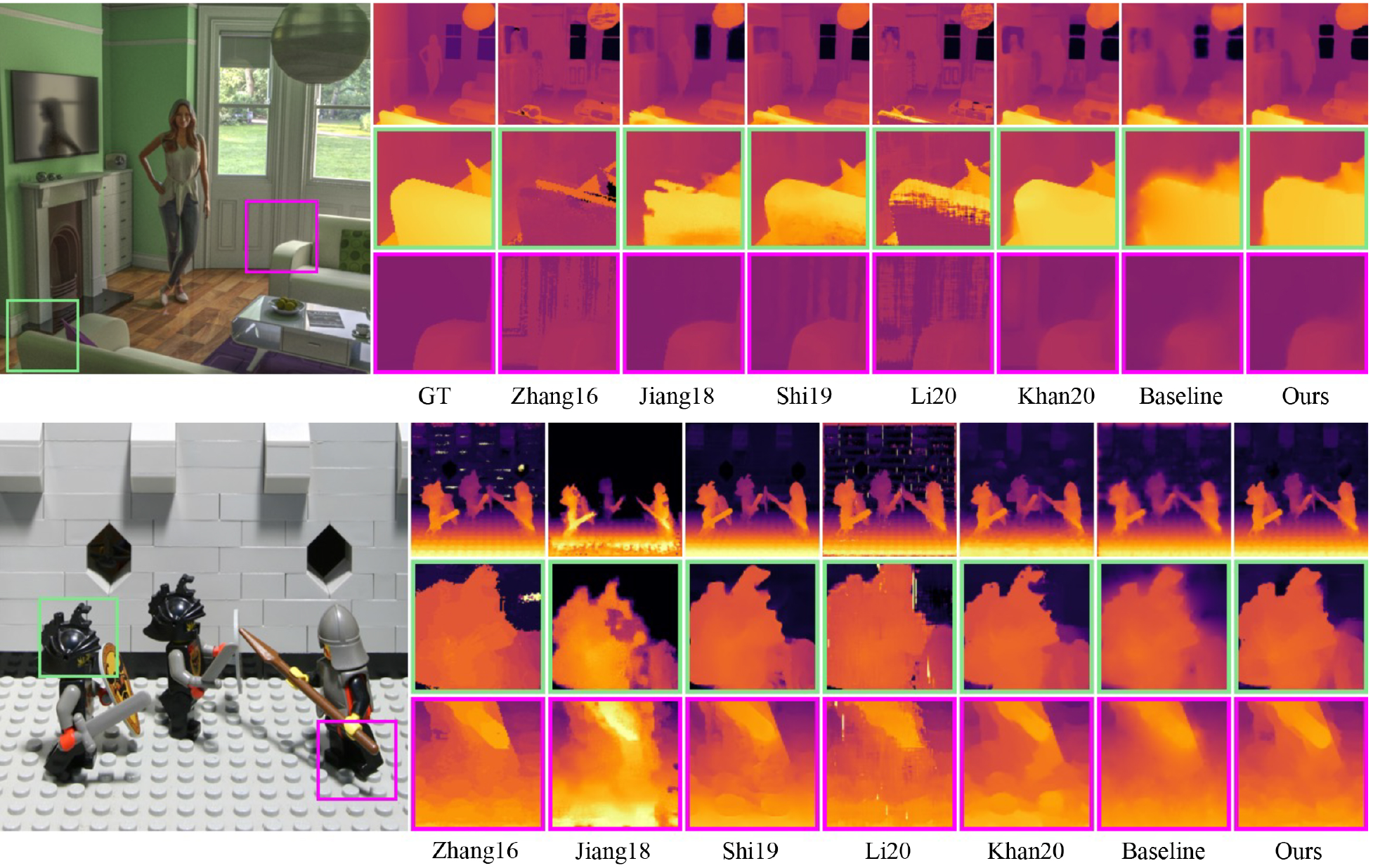}
\vspace{-0.25cm}
\caption{ \emph{Top:} Disparity results on the synthetic \emph{Living Room} light field. \emph{Bottom:} Disparity results on a real light field. \emph{Left to right}: Zhang et al.~\cite{zhang2016}, Jiang et al.~\cite{jiang2018}, Shi et al.~\cite{shi2019}, Li et al.~\cite{li2020}, Khan et al.~\cite{khan2020}, a baseline diffusion result without any optimization, our results and finally, for the top light field, ground truth.}
\label{fig:results}
\vspace{-2.5mm}
\end{figure*}

\mparagraph{Results}
\label{sec:lf-results}
While learning-based methods~\cite{jiang2018, shi2019, li2020} tend to do well on the HCI dataset, their quantitative performance degrades on the more difficult \emph{Piano} and \emph{Living Room} scenes (Tab.~\ref{table:quantitative-lf}). A similar qualitative trend shows the learning-based methods performing worse than diffusion on the real-world light fields (Fig.~\ref{fig:results}). Our method provides more consistent overall performance on all datasets. Moreover, the existing diffusion-based method~\cite{khan2020} has few pixels with very large errors but many pixels with small errors, producing consistently low MSE but more bad pixels. In contrast, our method consistently places in the top-three on the bad pixel metrics. 
We show additional results and error maps in our supplemental material. 
Finally, as is common, it is possible to post-process our results with a weighted median filter to reduce MSE (e.g., \emph{Dino} 0.54 vs.~0.86) at the expense of increased bad pixels (BP(0.01) of 39.6 vs.~25.6).

\subsection{Multi-view Stereo}
\mparagraph{Datasets} 
We path trace \textit{Living Room-MVS} and \textit{Piano-MVS} datasets at 512$\times$512. Each scene has five unstructured views with a mean baseline of approximately 25cm.

\mparagraph{Baselines and Metrics} We compare to dense reconstruction from COLMAP~\cite{schoenberger2016mvs} and to DeepMVS~\cite{huang18}.
Out method uses the sparse output of COLMAP as the initial point set, which is considerably sparser than the initial set for light fields (500 vs.~50k). To increase the number of points, we diffuse a preliminary depth map and optimize the smoothness parameter for 50 iterations. Then, we sample this result at RGB edges. Using this augmented set, we optimize all parameters in turns of 25 iterations, repeated 5 times. In addition, we also evaluate a variant of our method, Ours-C, with sparse labels initialized from the dense COLMAP output at RGB edges.

For metrics, we again use MSE, and also report the 25th percentile of absolute error as Q25. 
As the depth output of each method is ambiguous up to a scale, we estimate a scale factor for each result using a least squares fit to the ground truth at 500 randomly sampled valid depth pixels.

{
\newcolumntype{x}[1]{%
>{\centering\hspace{0pt}}p{#1}}%

\definecolor{best1}{rgb}{1, 0.8, 0} 
\definecolor{best2}{rgb}{0.78, 0.78, 0.78}
\definecolor{best3}{rgb}{0.79, 0.65, 0.44}

\newcommand{\gold}[1]{\colorbox{best1}{#1}}
\newcommand{\silver}[1]{\colorbox{best2}{#1}}
\newcommand{\bronze}[1]{\colorbox{best3}{#1}}

\setlength{\tabcolsep}{0.9pt}
\renewcommand{\arraystretch}{1.1}
\begin{table}[t]
\begin{center}
\vspace{-1mm}
\resizebox{\linewidth}{!}{%
\begin{tabular}{l|x{3.5em}|x{3.0em}|x{2.5em}|x{3.0em}||x{3.5em}|x{3.0em}|x{2.5em}|c}
\hline
\multirow{2}{*}{} & \multicolumn{4}{c}{MSE} & \multicolumn{4}{c}{Q25} \\
\cline{2-9}
& D-MVS & C-Map & Ours & Ours-C & D-MVS & C-Map & Ours &~Ours-C\\
\hline
\hline
\textit{Living Room-MVS}~& 1.99 & \bronze{1.37} & \silver{0.30} & \gold{0.17} & 64.9 & \silver{4.44} & \bronze{14.8} & \gold{4.22} \\
\textit{Piano-MVS}~& \bronze{1.51} & 2.56 & \silver{0.81} & \gold{0.69} & \bronze{6.87} & 42.6 & \silver{2.15} & \gold{1.37} \\
\hline
\textit{Average}& \bronze{1.75} & 1.97 & \silver{0.56} & \gold{0.43} & 35.9 & \bronze{23.5} & \silver{8.48} & \gold{2.80} \\
\hline
\end{tabular}
}
\end{center}
\vspace{-0.6cm}
\caption{Quantitative results for wider-baseline unstructured five-camera cases, as the \emph{Living Room-MVS} and \emph{Piano-MVS} scenes.}
\label{table:quantitative-mvs}
\vspace{-4mm}
\end{table}
\setlength{\tabcolsep}{1.4pt}
}

{
\newcolumntype{x}[1]{%
>{\centering\hspace{0pt}}p{#1}}%

\definecolor{best1}{rgb}{1, 0.8, 0} 
\definecolor{best2}{rgb}{0.78, 0.78, 0.78}
\definecolor{best3}{rgb}{0.79, 0.65, 0.44}

\newcommand{\gold}[1]{\colorbox{best1}{#1}}
\newcommand{\silver}[1]{\colorbox{best2}{#1}}
\newcommand{\bronze}[1]{\colorbox{best3}{#1}}

\setlength{\tabcolsep}{0.9pt}
\renewcommand{\arraystretch}{1.2}
\begin{table*}[htbp]
\begin{center}
\vspace{1mm}
\resizebox{\textwidth}{!}{%
\begin{tabular}{l|c|c|c|c|c|c||c|c|c|c|c|c||c|c|c|c|c|c||c|c|c|c|c|c}
\hline
\multirow{2}{*}{Light Field} & \multicolumn{6}{c}{MSE * 100} & \multicolumn{6}{c}{BP(0.1)} & \multicolumn{6}{c}{BP(0.3)} & \multicolumn{6}{c}{BP(0.7)}\\
\cline{2-25}
& \cite{zhang2016} & \cite{li2020} & \cite{jiang2018} & \cite{shi2019} & \cite{khan2020} & Ours & \cite{zhang2016} & \cite{li2020} & \cite{jiang2018} & \cite{shi2019} & \cite{khan2020} & Ours & \cite{zhang2016} & \cite{li2020} & \cite{jiang2018} & \cite{shi2019} & \cite{khan2020} & Ours & \cite{zhang2016} & \cite{li2020} & \cite{jiang2018} & \cite{shi2019} & \cite{khan2020} & Ours\\
\hline
\hline
\textit{Living Room~} & 0.67 & 0.57 & \silver{0.23} & \bronze{0.25} & \bronze{0.25} & \gold{0.20} & 59.5 & 58.5 & \silver{37.2} & 48.0 & \bronze{47.2} & \gold{30.3} & 43.3 & 42.7 & \silver{23.7} & 26.5 & \bronze{25.0} & \gold{17.5} & {17.0} & 16.6 & \bronze{11.4} & \silver{10.8} & {11.5} & \gold{9.23}\\
\textit{Piano} & {26.7} & 13.7 & {14.4} & \gold{8.66} & \bronze{12.7} & \silver{8.71} & {36.7} & {27.5} & \silver{24.7} & \bronze{27.0} & {37.6} & \gold{17.0} & 25.0 & 17.6 & \bronze{13.6} & \silver{11.4} & 20.0 & \gold{7.93} & {5.33} & \silver{4.13} & {5.88} & \bronze{4.29} & {4.95} & \gold{3.49} \\
\hline
\multirow{2}{*}{Light Field} & \multicolumn{6}{c}{MSE * 100} & \multicolumn{6}{c}{BP(0.01)} & \multicolumn{6}{c}{BP(0.03)} & \multicolumn{6}{c}{BP(0.07)}\\
\cline{2-25}
& \cite{zhang2016} & \cite{li2020} & \cite{jiang2018} & \cite{shi2019} & \cite{khan2020} &  Ours & \cite{zhang2016} & \cite{li2020} & \cite{jiang2018} & \cite{shi2019} & \cite{khan2020} & Ours & \cite{zhang2016} & \cite{li2020} & \cite{jiang2018} & \cite{shi2019} & \cite{khan2020} & Ours & \cite{zhang2016} & \cite{li2020} & \cite{jiang2018} & \cite{shi2019} & \cite{khan2020} & Ours\\
\hline
\hline
\textit{Sideboard} & \silver{1.02} & {1.89} & 1.96 & \bronze{1.12} &  \gold{0.89} & 2.23 & 78.0 & {62.3} & \silver{47.4} & \bronze{53.0} & 73.8 & \gold{43.0} & 42.0 & \silver{18.0} & \bronze{18.3} & {20.4} & 37.4 & \gold{16.5} & 14.4 & \gold{6.50} & {9.31} & \bronze{9.02} &  16.2 & \silver{8.35} \\
\textit{Dino} & \gold{0.41} & 3.28 & \bronze{0.47} & \silver{0.43} & 0.45 & 0.86 & 81.2 & 52.7 & \silver{29.8} & \bronze{43.0} & 69.4 & \gold{25.6} & 48.9 & \bronze{12.8} & \silver{8.81} & 13.1 & 30.9 & \gold{7.69} & 7.52 & 5.82 & \gold{3.59} & \bronze{4.32} & 10.4 & \silver{4.06}\\
\textit{Cotton} & 1.81 & 1.95 & \bronze{0.97} & \silver{0.88} & \gold{0.68} & 3.07 & 75.4 & 58.8 & \gold{25.4} & \bronze{38.6} & 56.2 & \silver{31.1} & 34.8 & 14.0 & \gold{6.30} & \bronze{9.60} & 18.0 & \silver{7.82} & 4.35 & \bronze{4.11} & \gold{2.02} & \silver{2.74} & 4.86 & \bronze{4.06}\\
\textit{Boxes} & \bronze{7.90} & \gold{4.67} & 11.6 & 8.48 & \silver{6.69} & 9.17 & 84.7 & 68.3 & \gold{51.8} & \bronze{66.5} & 76.8 & \silver{60.3} & 55.3 & \silver{28.0} & \gold{27.0} & 37.2 & 47.9 & \bronze{32.7} & \bronze{18.9} & \gold{13.4} & \silver{18.3} & 21.9 & 28.3 & 20.5\\
\hline
\end{tabular}
}
\end{center}
\vspace{-0.6cm}
\caption{ \textbf{(Best viewed in color)}~Quantitative comparison on synthetic light fields. The top three results are highlighted in \gold{gold}, \silver{silver} and \bronze{bronze}. BP($x$) is the number of \emph{bad pixels} which fall above threshold $x$ in error. Higher BP thresholds are used for \textit{Living Room} and \textit{Piano} as their average error is larger for all methods: they contain specular surfaces, larger depth ranges, and path tracing noise.}
\label{table:quantitative-lf}
\end{table*}
\setlength{\tabcolsep}{1.4pt}
}

\mparagraph{Results}
To account for the error in least squares, Table~\ref{table:quantitative-mvs} presents the minimum of ten different fits for each method. Both DeepMVS and COLMAP generate results with many invalid pixels. We assign such pixels the mean GT depth. Our method outperforms the baselines with a sparse point set ($\approx 700$ points) and generates smooth results by design that qualitatively have fewer artifacts (Fig.~\ref{fig:results-mvs}). Using 4$\times$ as many initial points ($\approx 2,800$ points) in the Our-C variant leads to additional improvements.


\section{Conclusion}
\label{sec:conclusion}

We present a method to differentiably render and diffuse a sparse depth point set such that we can directly optimize dense depth map reconstruction to minimize a multi-view RGB reprojection loss. 
While we recover depth maps, our approach can be interpreted as a point denoiser for diffusion, as related to volume rendering via radiative transfer, or as a kind of differentiable depth-image-based rendering.
We discuss higher-order weighting term design choices that make this possible, demonstrate our method's ability to reduce error in bad pixels, and discuss why remaining errors are difficult to optimize via reprojection from depth maps. 
In comparisons to both image processing and deep learning baselines, our method shows competitive performance, especially in reducing bad pixels.


\mparagraph{Acknowledgements} We thank the reviewers for their detailed feedback. Numair Khan thanks an Andy van Dam PhD Fellowship, and Min H.~Kim acknowledges the support of Korea NRF grant (2019R1A2C3007229).

\clearpage

{\small
\bibliographystyle{cvpr2021kit/ieee_fullname}
\bibliography{bibliography.bib}
}

\end{document}